%% file: main.tex
\documentclass{article}

\usepackage[preprint]{neurips_2022}

\usepackage[utf8]{inputenc} 
\usepackage[T1]{fontenc}    
\usepackage{hyperref}       
\usepackage{url}            
\usepackage{booktabs}       
\usepackage{amsfonts}       
\usepackage{nicefrac}       
\usepackage{microtype}      
\usepackage{xcolor}         

\usepackage{graphicx}
\usepackage{subfigure}
\usepackage{multirow}

\hypersetup{
    colorlinks=true,
    linkcolor=red,
    citecolor=cyan,
    filecolor=magenta,      
    urlcolor=cyan,
    }

\usepackage{amsmath}
\usepackage{amssymb}
\usepackage{mathtools}
\usepackage{amsthm}
\usepackage{bbm}
\usepackage{balance}
\usepackage{makecell}
\usepackage{sidecap}
\usepackage{tabulary}
\usepackage{colortbl}
\usepackage{bbding}

\newcommand{\ie}{\textit{i.e.}}
\newcommand{\eg}{\textit{e.g.}}
\newcommand{\argmax}{\mathop{\mathrm{argmax}}}

\newtheorem{theorem}{Theorem}

\title{Reliable Label Correction is a Good Booster When Learning with Extremely Noisy Labels}

\author{
  Kai Wang$^{1}$\thanks{Equal contribution. (kai.wang@comp.nus.edu.sg, xiangyupeng@comp.nus.edu.sg)},\; Xiangyu Peng$^{1}$\footnotemark[1],\; Shuo Yang$^{2}$,\; Jianfei Yang$^{3}$,\;\\ \textbf{Zheng Zhu}$^{4}$,\; \textbf{Xinchao Wang}$^{1}$,\;  \textbf{Yang You}$^{1}$\thanks{Corresponding author (youy@comp.nus.edu.sg).}
   \\
  $^{1}$National University of Singapore \quad 
  $^{2}$ University of Technology Sydney \\
  $^{3}$Nanyang Technological University, Singapore \quad
  $^{4}$Tsinghua University \\
Code: \url{https://github.com/xyupeng/LC-Booster}
}

\begin{document}

\maketitle
\input{sections/1.abstract}
\input{sections/2.intro}
\input{sections/3.preliminary}
\input{sections/4.method}
\input{sections/5.experiments}
\input{sections/6.conclusion}

\clearpage
\bibliographystyle{splncs04}
\bibliography{references}




\end{document}


\maketitle


\section{Proof of Theorem 1} \label{sec:proof1}
\textit{Proof:}
$\forall{\boldsymbol{x}} \in \mathcal{X}$, ${\tilde{\alpha}_{c}(\mathbf{x})}$ can be rewritten as:
\begin{align}
{\tilde{\alpha}_{c}(\mathbf{x})} =& P(\tilde{Y}=e_c, Y=e_c \mid \mathbf{x}) + P(\tilde{Y}=e_c, Y \neq e_c \mid \mathbf{x}) \nonumber \\
=& P(\tilde{Y}=e_c \mid Y=e_c) P(Y=e_c \mid \mathbf{x}) \nonumber \\
&+P(\tilde{Y}=e_c \mid Y \neq e_c) P(Y \neq e_c \mid \mathbf{x}) \nonumber \\
=& P(\tilde{Y}=e_c \mid Y=e_c) {{\alpha}_{c}(\mathbf{x})} + \rho_{c} (1 - {{\alpha}_{c}(\mathbf{x})}) \nonumber \\
\leq& {{\alpha}_{c}(\mathbf{x})} + \rho_{c} (1 - {{\alpha}_{c}(\mathbf{x})}) \nonumber \\
=& (1-\rho_{c}){{\alpha}_{c}(\mathbf{x})}+\rho_{c}
\end{align}
If ${{\alpha}_{c}(\mathbf{x})} \leq \frac{1}{2}$, we have

\begin{align}
{\tilde{\alpha}_{c}(\mathbf{x})}
\leq& (1-\rho_{c}){{\alpha}_{c}(\mathbf{x})}+\rho_{c} \nonumber \\
\leq& \frac{1}{2}(1-\rho_{c})+\rho_{c} \nonumber \\
=&  \frac{1+\rho_{c}}{2} \nonumber \\
\end{align}
and its contrapositive
 
 \begin{equation}
{\tilde{\alpha}_{c}(\mathbf{x})}>  \frac{1+\rho_{c}}{2}  \Longrightarrow  \alpha_{c}(\mathbf{x}) > \frac{1}{2}.
\end{equation}

\section{Additional Training Details} \label{sec:detail}
For CIFAR-10 and CIFAR-100, we perform ReCo at the 100th epoch and tune $\tau_{ps}$ for different noise rates. We show the values for $\tau_{ps}$ and $\lambda_u$ in Tab.~\ref{tab:tau_ps}. We use smaller $\tau_{ps}$ for larger noise or more classes, as training with either of them decreases the confidence of the model. We set $\lambda_r$ as 1 across all experiments. Choice of $\lambda_u$ and $\lambda_r$ follows \cite{li2020dividemix}. Experiments on CIFAR-10 and CIFAR-100 are done using 2 RTX 3090 GPUs.
\input{tables/tau_ps}

For Clothing1M and WebVision, we use the same hyper-parameters $\tau_{ps}=0.8$, $\lambda_u=0$ and $\lambda_r=1$. We train the model for 100 epochs on both datasets, and perform ReCo at the 60th epochs. The warm up period is 1 epoch. We use SGD optimizer with a momentum of 0.8, a weight decay of 0.001, and a batch size of 32. For Clothing1M, the initial learning rate is set as 0.002 and reduced by a factor of 10 after 50 epochs. For WebVision, the initial learning rate is set as 0.01 and reduced by a factor of 10 after 50 epochs. Experiments on Clothing1M and WebVision are done using 8 RTX 3090 GPUs.

\section{More Visualization} \label{sec:comp_vis}
\begin{figure}[htp] 
    \centering
    \subfigure[DivideMix]
    {\label{fig:tsne_dm_10c}{\includegraphics[width=0.48\textwidth, trim=0cm 0cm 0.5cm 0.3cm, clip]{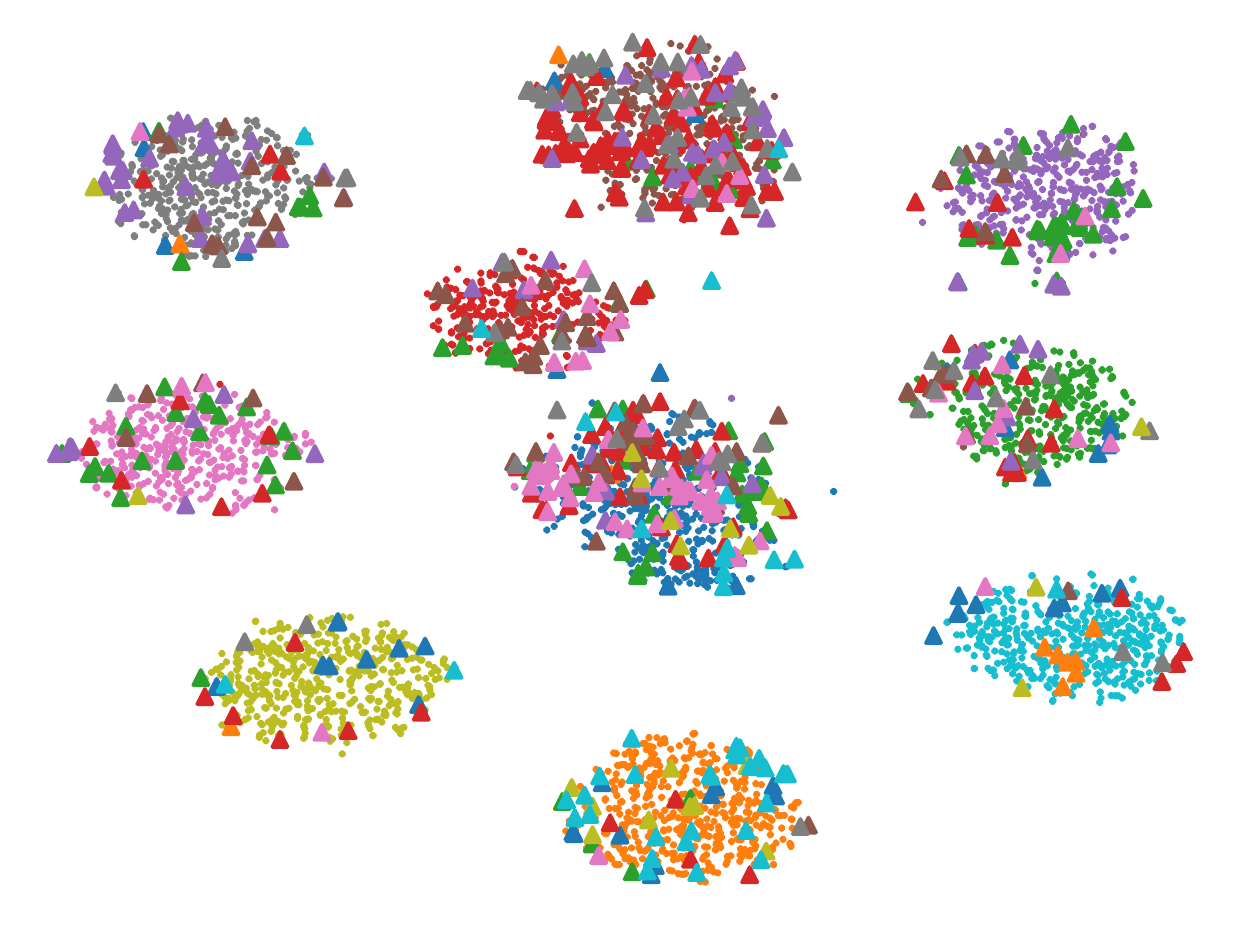}}}
    \subfigure[LC-Booster]
    {\label{fig:tsne_lc_10c}\includegraphics[width=0.48\textwidth, trim=0cm 0cm 0.5cm 0.3cm, clip]{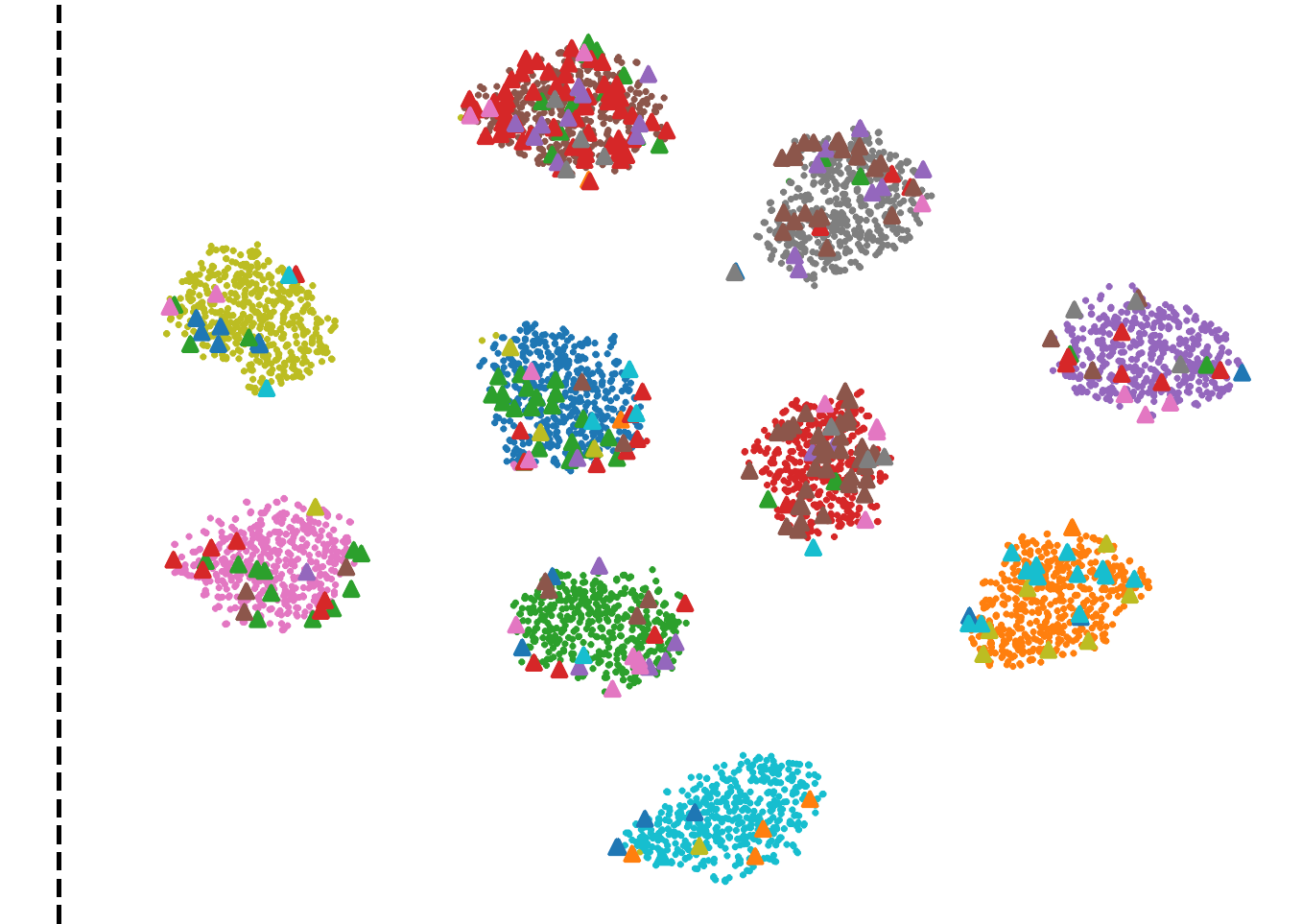}}
\caption{Visualization of embedded features on CIFAR-10 with 90\% symmetric noise. 10 classes are visualized in different colors. Correct predictions are marked as circles and false predictions as triangles. Best viewed in color.}
\label{fig:tsne_10c}
\end{figure}

We show a complete distribution of all 10 classes of CIFAR-10 in Fig.~\ref{fig:tsne_10c}. It can be seen from the figure that clusters learned by LC-Booster are more compact. Moreover, our method has fewer false predictions (marked as triangles) compared with DivideMix, showing its robustness under extreme label noise.

\begin{figure}[htp] 
    \centering
    \subfigure[]
    {\label{fig:auc_ep50}{\includegraphics[width=0.4\textwidth]{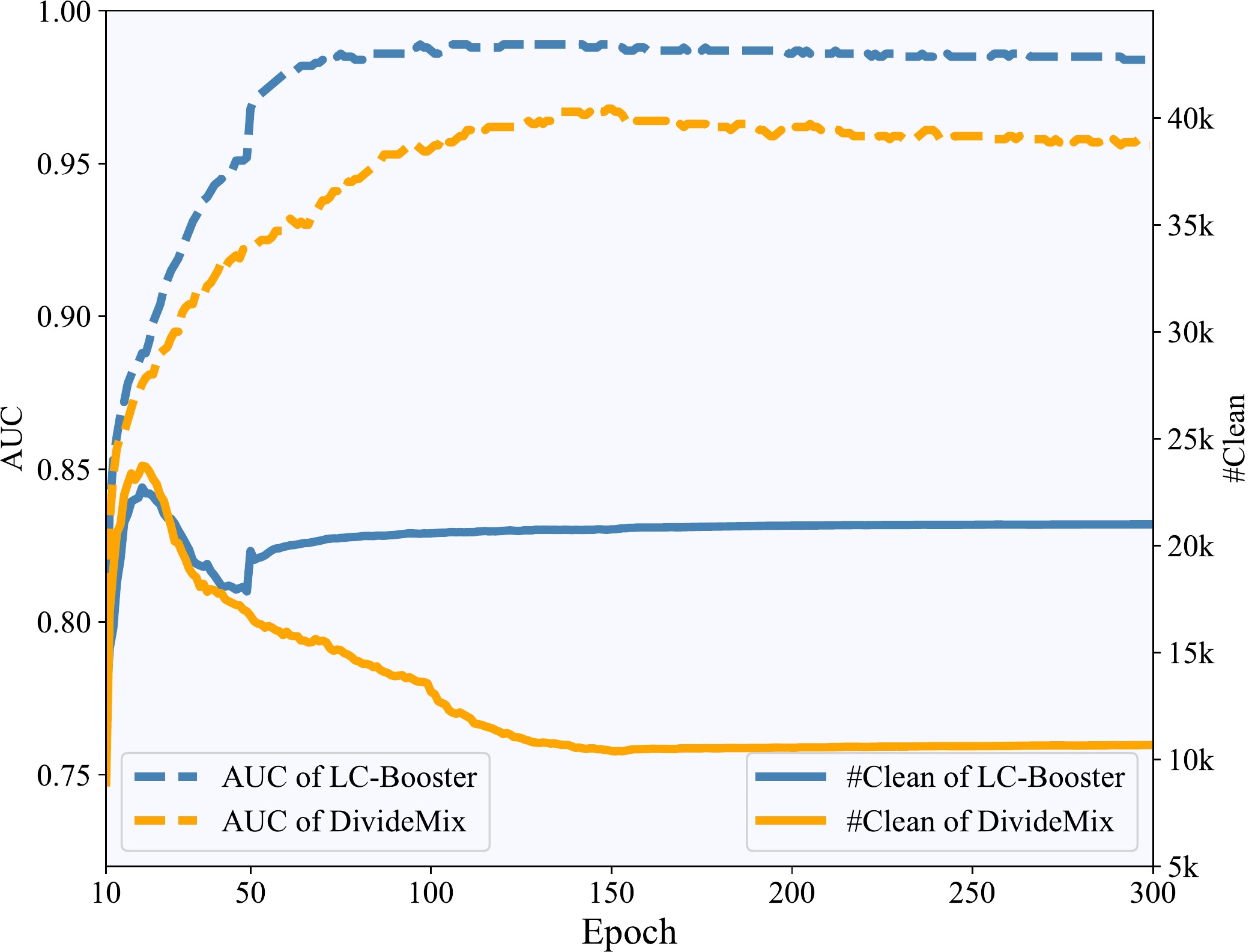}}}
    \subfigure[]
    {\label{fig:auc_ep100}\includegraphics[width=0.4\textwidth]{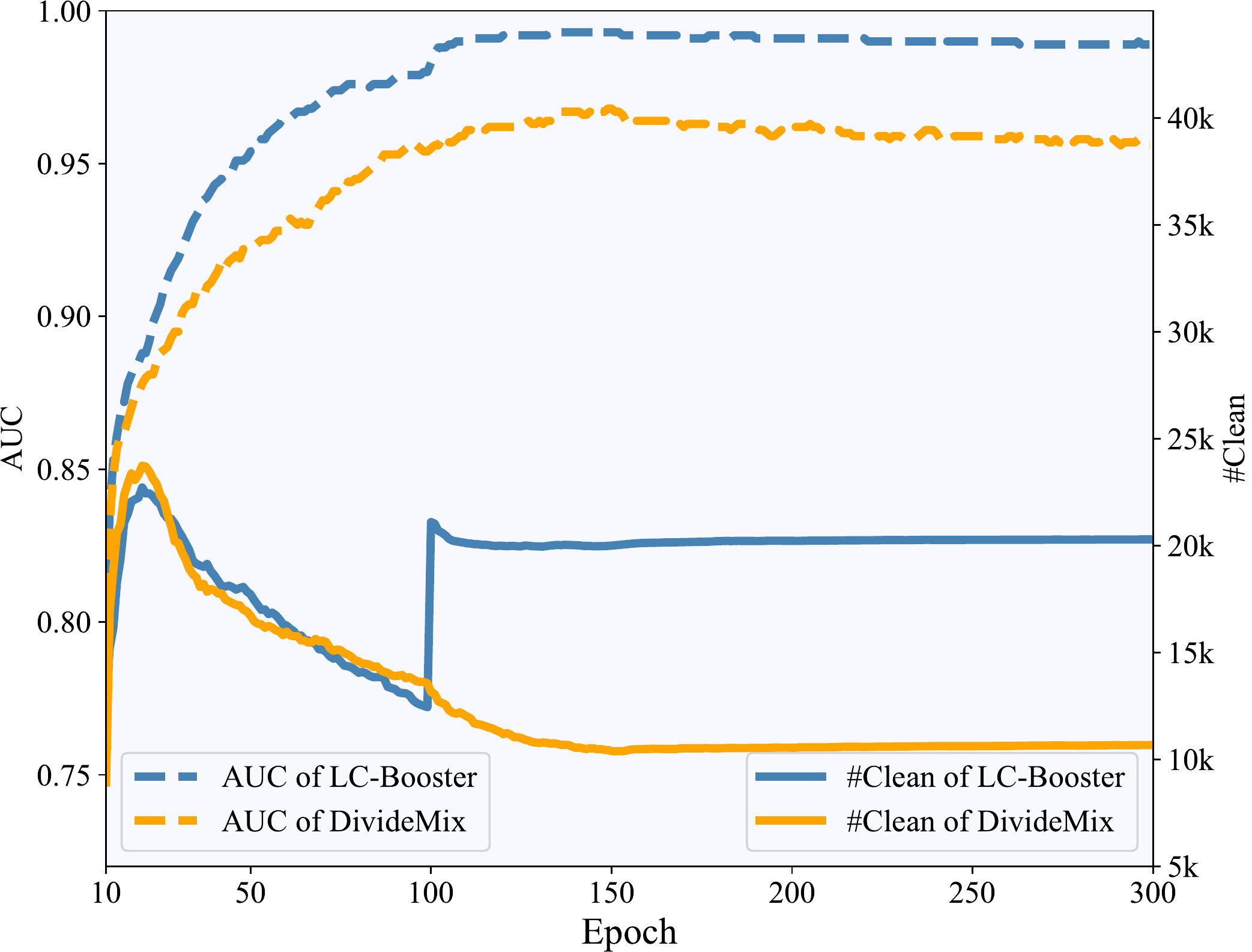}}
    \subfigure[]
    {\label{fig:auc_ep150}\includegraphics[width=0.4\textwidth]{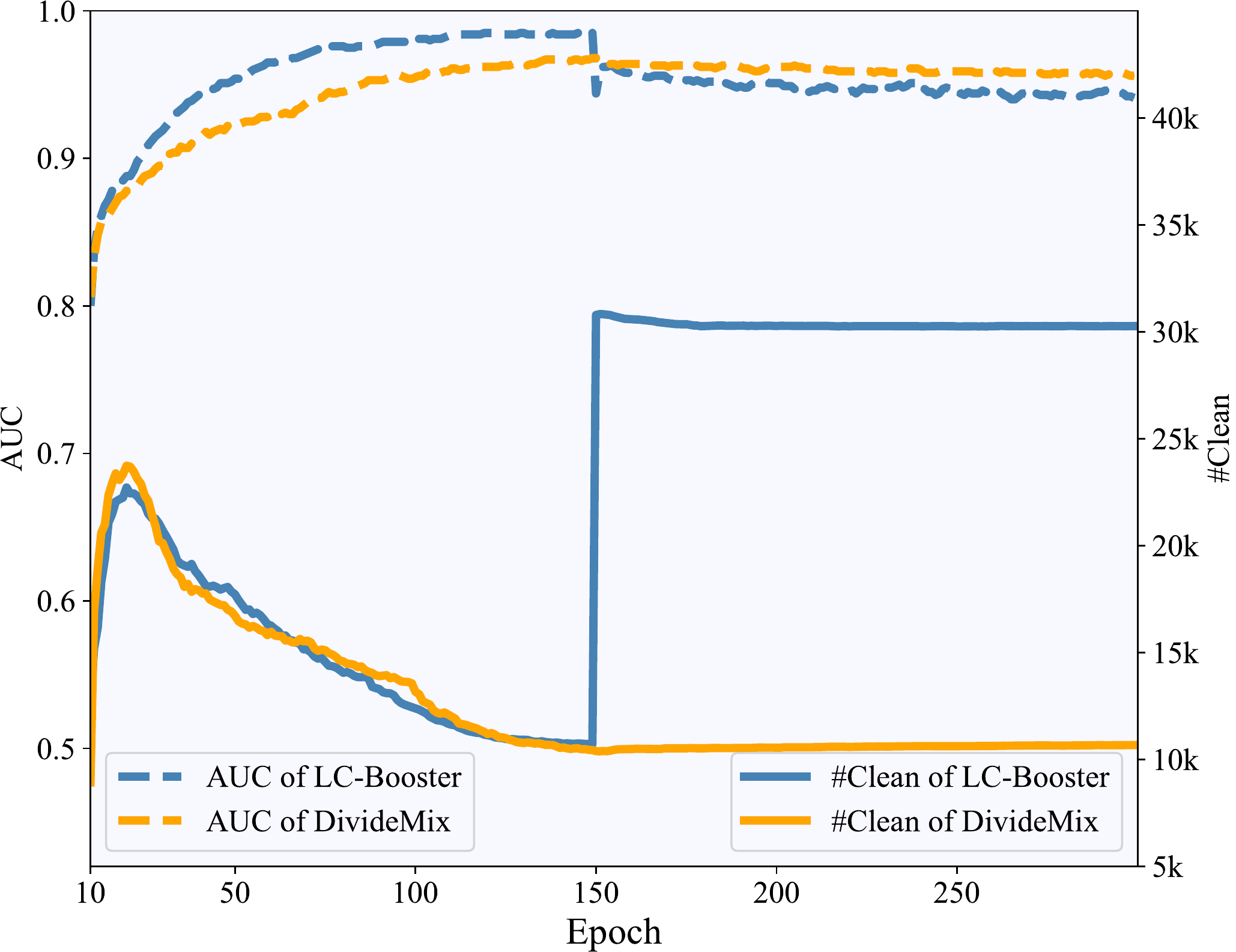}}
    \subfigure[]
    {\label{fig:auc_ep200}\includegraphics[width=0.4\textwidth]{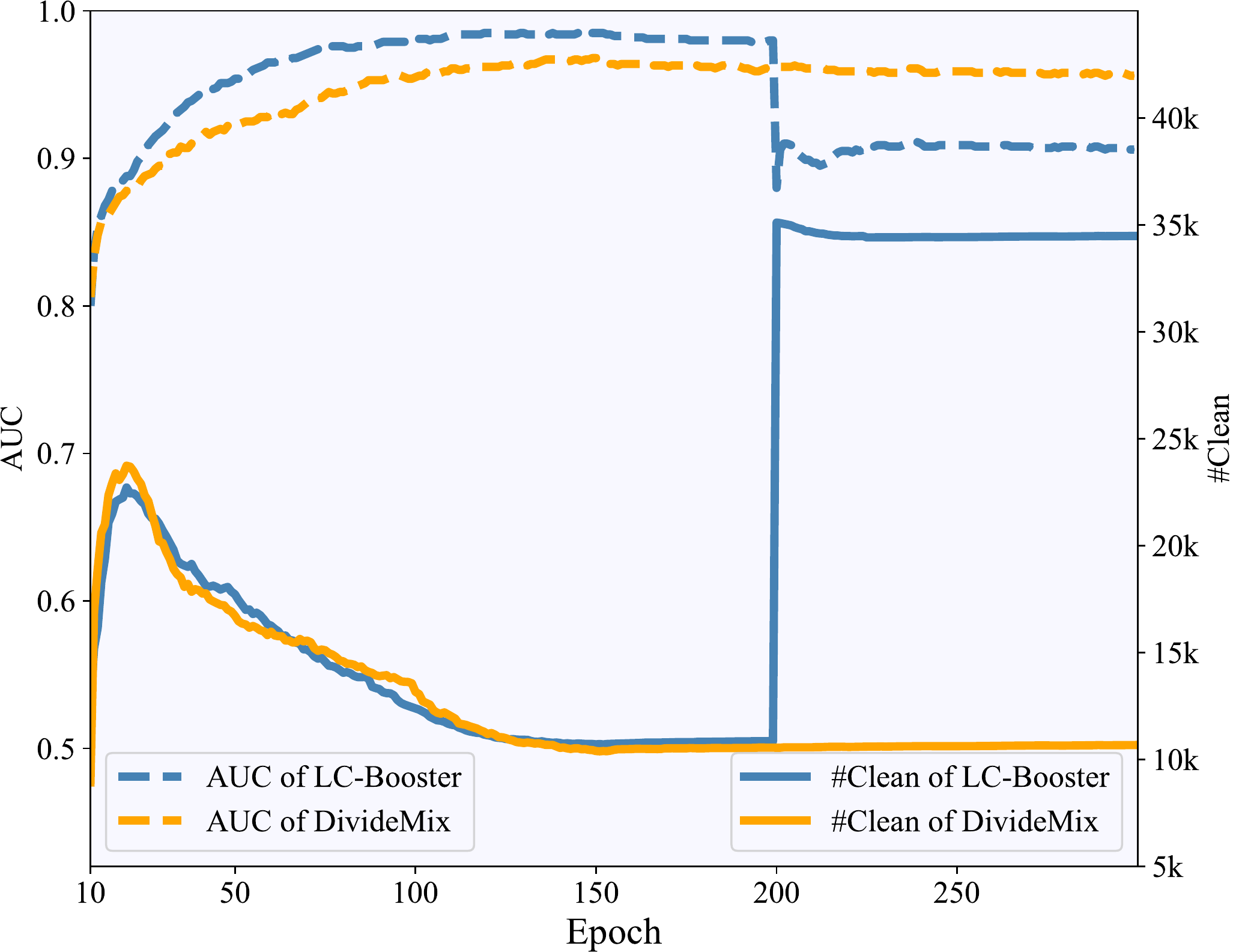}}
\caption{Curves of AUC and size of clean set (\#Clean) on CIFAR-10 with 90\% symmetric noise. Higher AUC indicates that clean samples are selected more precisely based on GMM. (a), (b), (c), (d) show the curves of performing ReCo at the 50th, 100th, 150th and 200th epoch, respectively.}
\label{fig:more_auc}
\end{figure}

We also show curves of AUC and size of clean set for different re-labeling epochs in Fig.~\ref{fig:more_auc}. Re-labeling at the 100th achieves the highest AUC and the largest size. Earlier re-labeling at the 50th epoch may not be very much reliable, since the model has not been trained sufficiently. Re-labeling in later epochs (\eg, the 150th and 200th epoch) results in a larger expansion of the clean set but a drop in AUC. We hypothesize this is due to overfitting to noisy samples.

\clearpage
\bibliographystyle{unsrt}
\bibliography{references}

%% file: sections/1.abstract.tex
\begin{abstract}
    Learning with noisy labels has aroused much research interest since data annotations, especially for large-scale datasets, may be inevitably imperfect. Recent approaches resort to a semi-supervised learning problem by dividing training samples into clean and noisy sets. This paradigm, however, is prone to significant degeneration under heavy label noise, as the number of clean samples is too small for conventional methods to behave well. In this paper, we introduce a novel framework, termed as LC-Booster, to explicitly tackle learning under extreme noise. The core idea of LC-Booster is to incorporate label correction into the sample selection, so that more purified samples, through the reliable label correction, can be utilized for training, thereby alleviating the confirmation bias. Experiments show that LC-Booster advances state-of-the-art results on several noisy-label benchmarks, including CIFAR-10, CIFAR-100, Clothing1M and WebVision. Remarkably, under the extreme 90\% noise ratio, LC-Booster achieves 92.9\% and 48.4\% accuracy on CIFAR-10 and CIFAR-100, surpassing state-of-the-art methods by a large margin.
\end{abstract}

%% file: sections/2.intro.tex
\section{Introduction}

Contemporary large-scale datasets are prone to be contaminated by noisy labels, due to inevitable human failure, unreliable open-source tags~\cite{mahajan2018exploring}, challenging labeling tasks~\cite{frenay2013classification}, and errors made by machine generation~\cite{kuznetsova2020open}. Training deep neural networks (DNNs) with a non-trivial amount of label noise could result in poor generalization performance~\cite{zhang2017understanding}. This behavior can be explained by the over-parameterization characteristics of DNN~\cite{allen2018learning} and the consequent strong memorization ability~\cite{arpit2017closer}. 

Recently, a variety of approaches have been proposed to train robust DNNs in a noisy label environment. Some of the works adopt label correction to revise noisy labels based on network predictions~\cite{wang2020suppressing, wu2020learning}, thus reducing the noise level in the dataset. However, these methods may suffer from confirmation bias~\cite{arazo2020pseudo}, which refers to the accumulation of prediction errors in the process of re-labeling. More recently, a series of works based on sample selection (SS) stand out and show promising results. The main idea is to distill clean samples from noisy data so that the negative influence of label noise could be mitigated. Among these methods, Co-teaching~\cite{han2018co} and Co-teaching+~\cite{yu2019does} select a portion of small-loss instances as clean samples since DNNs tend to learn easy and clean samples first before overfitting to noisy labels~\cite{arpit2017closer}. Another representative work DivideMix~\cite{li2020dividemix} fits a Gaussian Mixture Model (GMM) on the per-sample loss distribution for adaptive sample selection. To avoid confirmation bias, it adopts a two-network structure where the GMM for one network is used to divide training data for the other network. 

\begin{figure}[h]
\centering
\begin{minipage}{.55\textwidth}
    \includegraphics[width=\textwidth]{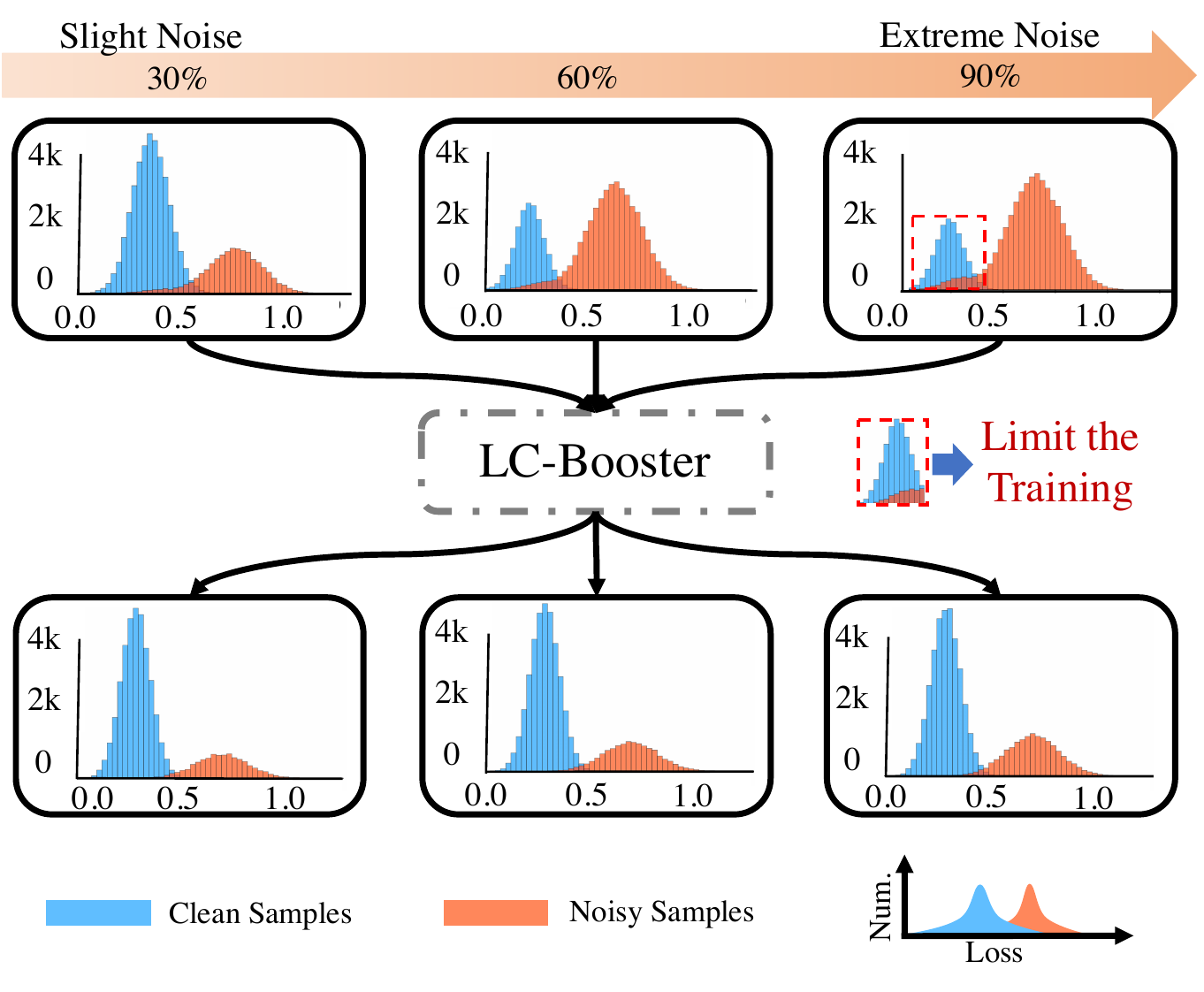}
    \label{fig:motivation}
\end{minipage}\hspace{2mm}
\begin{minipage}{.4\textwidth}\vspace{-1.2em}
    \caption{The motivation of the proposed LC-Booster. From left to right, the noise ratio increases from 30\% to 90\%. The histograms show normalized losses used to divide clean and noisy sets based on Gaussian Mixture Model (GMM). Conventionally, the number of clean samples shrinks significantly as noise ratio rises (the top row), which may limit the training of models. In this work, we find that noisy labels could be reliably revised in the sample selection setting (the bottom row), so that more purified clean samples could be involved in training to boost model performance.}
\end{minipage}
\vspace{-2.5em}
\end{figure}

Though SS-based methods can effectively pick out most of the clean samples, their performance would degenerate significantly when confronting extremely noisy labels. This can be easily understood, as the number of the clean samples is intrinsically limited by the noisy dataset. Training with such a small labeled set may lead to insufficiently trained models, and consequently, poor generalization performance. This naturally raises a question: Is it possible to enlarge the clean set for further performance-boosting, on top of filtering out adverse noisy samples? 

To answer this question, we first identify that sample selection based methods intrinsically lack a mechanism to produce new clean samples, despite their excellent ability to distill a much clean set from large label noise. This inspires us that an extra technique is necessary to achieve it. To this end, we rethink the feasibility of label correction, a conventional tool to turn noisy samples into clean ones, in the new setting of sample selection. Previous label correction methods, as mentioned above, are prone to suffer from confirmation bias, since model predictions could be severely hurt when heavy label noise is involved in training. However, in the SS setting, the problem of confirmation bias could be largely mitigated, as much label noise is filtered out and only a highly purified clean set is used for supervised training. Based on the trusted clean set, predictions of the model are much more reliable sources for label correction. In fact, attempts have been made to increase the reliability of label correction. \cite{reed2014training} use bootstrapping to generate new labels. ~\cite{zhang2020distilling} leverage a side clean set (\ie, clean samples given in advance) as anchors to reconstruct the noisy dataset. However, we argue that neither bootstrapping nor anchor clean samples are necessary, as in the SS setting a trusted clean set is naturally provided which label correction could rely on.

Based on this insight, we propose LC-Booster, a noise-robust framework that leverages label correction jointly with sample selection for a performance boost. In this framework, the clean set could keep a high label precision with adaptive sample selection while extending its size thanks to reliable label correction. Specifically, we start by warming up the model for a few iterations, so that some easy patterns can be learned first. Then, we divide clean and noisy samples based on GMM loss modeling as in~\cite{li2020dividemix}, where labels of the clean samples are kept for supervised loss and noisy samples are treated in an unsupervised manner. For better generalization, we also adopt a hybrid augmentation (H-Aug.) strategy that enforces consistency on both weak-weak and weak-strong augmented views. At around the middle stage, Reliable Label Correction (ReCo) is adopted to revise the labels for both clean and noisy samples. We theoretically show that the precision of revised labels can be guaranteed with a proper choice of threshold. With ReCo involved in training, the clean set can be improved in terms of both purity and quantity (shown at the bottom of Fig.~\ref{fig:motivation}), which could guide the model to learn better representations.

To the best of our knowledge, we are the first to leverage the strengths of both sample selection and label correction in a unified framework, despite the simplicity of the individual technique. We validate the effectiveness of LC-Booster on several noisy-label benchmarks, including CIFAR-10, CIFAR-100, Clothing1M, and WebVision. Our approach achieves state-of-the-art results on most of these datasets. Remarkably, under the extreme 90\% noise ratio, our approach achieves 92.9\% and 48.4\% accuracy on CIFAR-10 and CIFAR-100, surpassing the state of the art by 1.0\% and 7.2\% respectively. Our main contributions can be summarized as:
\begin{itemize}
    \item We find that label correction can be naturally leveraged with sample selection as a new paradigm for learning with noisy labels. The two techniques could work jointly to make a larger and more purified clean set.
    \item We propose LC-Booster, a simple yet efficient framework that could boost performance under (extreme) label noise. LC-Booster adopts H-Aug. for better generalization and ReCo for precisely revising labels with backing up theorem. 
    \item We experimentally show that LC-Booster advances state-of-the-art results on multiple benchmarks, especially under heavy label noise. We also conduct extensive ablation studies to illustrate the effects of our method. 
\end{itemize}

%% file: sections/3.preliminary.tex
\section{Preliminaries}

\subsection{Problem Formulation} \label{sec:problem_form}
In the problem of learning with noisy labels (LNL), we consider the noisy training set $\mathcal{S}=\{(\mathbf{x}_i, \tilde{\mathbf{y}}_i)\}_{i=1}^{N} = (\mathcal{S}_{\mathbf{x}}, \mathcal{S}_{\tilde{\mathbf{y}}})$, where $\mathbf{x}_i$ is the $i^{th}$ image and $\tilde{\mathbf{y}}_i \in \{0,1\}^C$ is the one-hot label over $C$ classes. $(\mathbf{x}_i, \tilde{\mathbf{y}}_i)$ is an image-target pair drawn from random variables $(X, \tilde{Y}) \sim (\mathcal{D}_X, \mathcal{D}_{\tilde{Y}})$, where $\mathcal{D}_X$ and $\mathcal{D}_{\tilde{Y}}$ denote the data distribution and the noisy label distribution, respectively. Similarly, we use $Y \sim \mathcal{D}_Y$ to represent the distribution for ground truth labels, which is unknown in the LNL problem setting. The noise rate of given class $c$ is defined as $\rho_c = P(\tilde{Y} = e_c | Y \neq e_c)$, with $e_c$ denoting the one-hot vector activated in position $c$, and the overall noise rate is $\rho = \frac{1}{C} \sum_{i=1}^{C} P(\tilde{Y} = e_i | Y \neq e_i)$. Generally, $\tilde{Y}$ can be 
divided into two types: 

\vspace{-1em}
\begin{itemize}
    \item Symmetric noise $\tilde{Y}_{sym}$. The label flips to a random class with a fixed probability $\eta$. With symmetric noise, we have $P(\tilde{Y}_{sym} = e_i | Y=e_i) = 1 - \eta + \frac{\eta}{C}$ and $P(\tilde{Y}_{sym} = e_j | Y=e_i) = \frac{\eta}{C}, \forall{i,j} \in \{1,2,...,C\}, i \neq j$. \\
    \vspace{-1em}
    \item Asymmetric noise $\tilde{Y}_{asym}$. The label flips to a certain class defined in a dictionary $\mathcal{M}$, which is built on the mapping between similar classes, \ie, \textit{cat} $\longrightarrow$ \textit{dog}, \textit{deer} $\longrightarrow$ \textit{horse}, \textit{bird} $\longrightarrow$ \textit{airplane}. With flipping probability $\eta$, we can arrive at $P(\tilde{Y}_{asym} = e_i | Y=e_i) = 1 - \eta + \mathbbm{1}_{\mathcal{M}(i)=i} \cdot \eta, \forall{i} \in \{1,2,...,C\}.$
\end{itemize}

\subsection{Background} \label{sec:background}
We consider sample selection methods~\cite{jiang2018mentornet, li2020dividemix} as the base of our approach, which have recently shown great promise in dealing with label noise. Typically, these methods divide training samples into the clean and noisy sets, denoted by $\mathcal{X}$ and $\mathcal{U}$ respectively. The labels of the clean set $\mathcal{X}$ are used for supervised training, since it has a higher label precision, while the noisy set $\mathcal{U}$ is treated unsupervised or simply abandoned due to its large noise ratio. A two-network structure is also commonly applied in state-of-the-art noise-robust models~\cite{li2020dividemix, cordeiro2021longremix}, where the two classifiers $f_{\theta_1}, f_{\theta_2}$ share the same structure but have different groups of parameters $\theta_1$, $\theta_2$. The training of $f_{\theta_1}$ and $f_{\theta_2}$ is performed in a \textit{co-teaching} manner~\cite{han2018co} (\ie, the division made by a network is used by the other), to mutually reduce prediction error and achieve a favorable ensemble effect.

Another important factor is how to precisely pick out clean samples. A dynamic selection approach is based on loss modeling, namely the small-loss strategy, leveraging the fact that DNNs tend to learn simple patterns first before overfitting to noisy labels~\cite{arpit2017closer}. In~\cite{arazo2019unsupervised} and ~\cite{li2020dividemix}, a clean probability is modeled for each sample as $P_i^{clean}(\ell_i, \{\ell_j\}_{j=1}^N, \gamma)$, with $\ell_i=-\sum_c \tilde{\mathbf{y}}_i^c \cdot \mathrm{log}(f_{\theta}^c(\mathbf{x}_i))$ being the classification loss for sample $i$ and $\gamma$ being the hyper-parameter.

In this work, we fit a two-component Gaussian Mixture Model (GMM)~\cite{permuter2006study} to the loss distribution as in~\cite{li2020dividemix}, and $P^{clean}$ is the posterior probability of the lower-mean component that fits small losses. In this way, we can write the clean and noisy set as
\begin{equation} \label{eq:cn1}
\begin{split}
    \mathcal{X} & = \{(\mathbf{x}_i, \tilde{\mathbf{y}}_i)|(\mathbf{x}_i, \tilde{\mathbf{y}}_i) \in \mathcal{S}, P_i^{clean} \ge \tau_c\}, \\
    \mathcal{U} & = \{(\mathbf{x}_i, p_i)|\mathbf{x}_i \in \mathcal{S}_{\mathbf{x}}, P_i^{clean} < \tau_c\},
\end{split}
\end{equation}
where $\tau_c$ is the probability threshold for the clean set and $p_i = \frac{1}{2}(f_{\theta_1}(\mathbf{x}_i) + f_{\theta_2}(\mathbf{x}_i))$ is the softmax probabilities predicted jointly by $f_{\theta_1}$ and $f_{\theta_2}$~\cite{li2020dividemix}.

After the division, the two classifiers $f_{\theta_1}$, $f_{\theta_2}$ are trained on $\tilde{\mathcal{X}}$ and $\tilde{\mathcal{U}}$ with a semi-supervised approach. Following~\cite{li2020dividemix}, we use MixMatch~\cite{berthelot2019mixmatch} to transform $\mathcal{X}$ and $\mathcal{U}$ into mixed clean and noisy sets $\mathcal{X}'$ and $\mathcal{U}'$, where
\begin{equation} \label{eq:mix_cn1}
\begin{split}
    \mathcal{X}' = \{ & (l(\mathbf{x}_i, \mathbf{x}_j, \lambda), l(\tilde{\mathbf{y}}_i, \mathbf{y}_j, \lambda)) | (\mathbf{x}_i, \tilde{\mathbf{y}}_i) \in \mathcal{X}, (\mathbf{x}_j, \mathbf{y}_j) \in \mathcal{X} \cup \mathcal{U}\}, \\
    \mathcal{U}' = \{ & (l(\mathbf{x}_i, \mathbf{x}_j, \lambda), l(p_i, \mathbf{y}_j, \lambda)) | (\mathbf{x}_i, p_i) \in \mathcal{U}, (\mathbf{x}_j, \mathbf{y}_j) \in \mathcal{X} \cup \mathcal{U}\},
\end{split}
\end{equation}
$l(\cdot,\cdot,\lambda)$ is the linear interpolation function (\ie, $l(\mathbf{x}_i, \mathbf{x}_j, \lambda)=\lambda \mathbf{x}_i + (1-\lambda) \mathbf{x}_j$), and $\lambda \sim \mathrm{Beta}(\alpha, \alpha)$ is a real number within $[0,1]$ sampled from a beta distribution. We make sure that $|\mathcal{X}| = |\mathcal{X}'|$ and $|\mathcal{U}| = |\mathcal{U}'|$. The training objective is to minimize
\begin{equation} \label{eq: VR1}
    \mathcal{L}_{\mathrm{VR}}(\mathcal{X}', \mathcal{U}') = \mathcal{L}_x(\mathcal{X}') + \lambda_u \mathcal{L}_u(\mathcal{X}'),
\end{equation}
where
\begin{equation} \label{eq:ce}
    \mathcal{L}_x(\mathcal{X}') = \frac{-1}{|\mathcal{X}'|}\sum_{\substack{(\mathbf{x}_i,\mathbf{y}'_i) \\ \in \mathcal{X}'}} \sum_c {\mathbf{y}'_i}^c \cdot \mathrm{log}(f_{\theta}^c(\mathbf{x}_i)), \quad
    \mathcal{L}_u(\mathcal{U}') = \frac{1}{|\mathcal{U}'|}\sum_{\substack{(\mathbf{x}_i,\mathbf{y}'_i) \\ \in \mathcal{U}'}} \|\mathbf{y}'_i-f_{\theta}(\mathbf{x}_i)\|_{2}^{2},
\end{equation}

and $\lambda_u$ controls the strength of the unsupervised loss. This objective is known as vicinal risk minimization (VRM), which is shown to be capable of reducing the memorization of corrupt labels~\cite{zhang2017mixup}.

%% file: sections/4.method.tex
\section {Methodology}

\subsection{Overview of the LC-Booster}
\begin{figure*}[htp]
    \centering
    \includegraphics[width=1.0\textwidth]{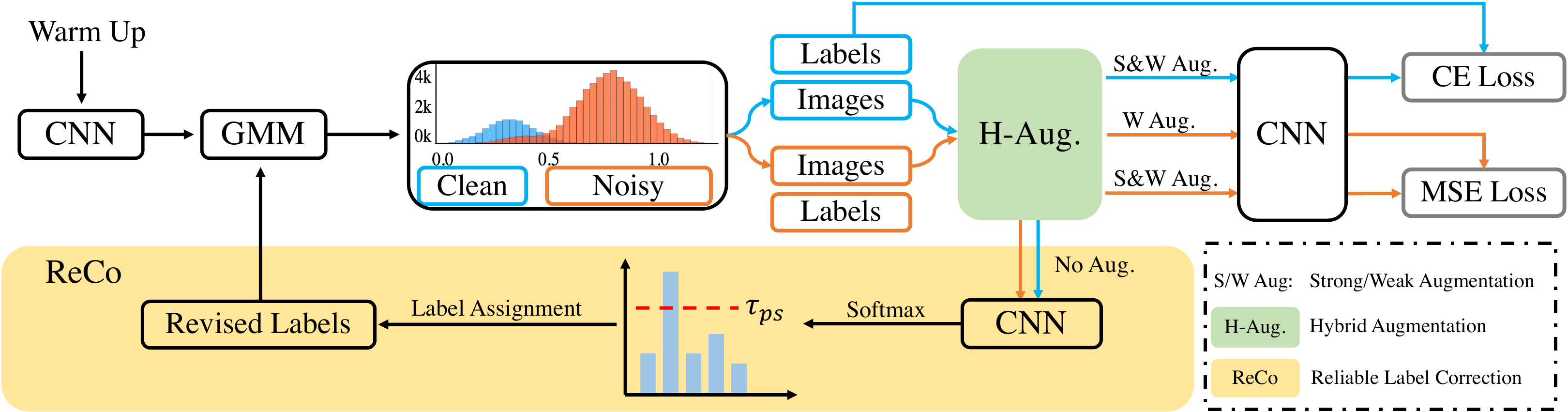}
    \caption{An overview of the proposed LC-Booster framework. We first warm up the model for a few iterations, and then fit a GMM to the loss distribution to separate clean or noisy sets. We then adopt H-Aug, which enforces consistency between weak-weak and weak-strong views. At the middle of training, we perform ReCo to revise the labels for all samples. The revised labels are used for GMM and CE Loss in the rest of training.}
    \label{fig:pipeline}
\end{figure*}

An overview of the LC-Booster framework is presented in Fig.~\ref{fig:pipeline}. We first warm up the model for a few iterations by training with all data, so that some easy patterns can be learned first. Then, we divide samples into clean and noisy sets $\mathcal{X}$, $\mathcal{U}$ defined in Eq.~\ref{eq:cn1}, and use MixMatch~\cite{berthelot2019mixmatch} to train the model. For better generalization, we adopt a hybrid augmentation (H-Aug.) strategy that transforms images into weakly and strongly augmented views. We use the labels to calculate Cross-Entropy Loss (CE Loss) for clean samples and minimize Mean Square Error Loss (MSE Loss) between weak-weak and weak-strong views of noisy samples. At the middle of training, Reliable Label Correction (ReCo) is adopted to revise the labels in both clean and noisy sets. The revised labels are then used in the rest of the training, which allows a larger clean set to be selected by GMM.

\subsection{Reliable Label Correction} \label{sec:reco}
Reliable Label Correction (ReCo) aims to deal with the meagre clean set problem in sample selection methods. With such a small labeled clean set, the generalization performance of a network could degenerate significantly, since DNNs are known to be data-hungry. To better leverage the noisy data, we propose to revise the labels in the training set $\mathcal{S}$ based on network predictions, so that more samples could be involved in the clean set with supervised signals. Specifically, we perform label correction by assigning those high confidence samples with hard pseudo labels, which are in accordance with their highest predictions. This gives us a new training set $\hat{\mathcal{S}}$ that mixes both raw and pseudo labels. Formally, it can be written as

\begin{small}
\begin{equation} \label{eq:S_hat}
\begin{gathered} 
    \hat{\mathcal{S}}_r = \{(\mathbf{x}_i, \tilde{\mathbf{y}}_i) | \forall{(\mathbf{x}_i, \tilde{\mathbf{y}}_i) \in \mathcal{S}} : \max_c p_i^c < \tau_{ps} \}, \\
    \hat{\mathcal{S}}_{ps} = \{(\mathbf{x}_i, e^k) | \forall{\mathbf{x}_i \in \mathcal{S}_{\mathbf{x}}} : \max_c p_i^c \ge \tau_{ps}, k=\argmax_c \mathbf{p}_i^c \}, \\ 
    \hat{\mathcal{S}} = \hat{\mathcal{S}}_r \cup \hat{\mathcal{S}}_{ps},
\end{gathered}
\end{equation}
\end{small}

where $p_i = \frac{1}{2}(f_{\theta_1}(\mathbf{x}_i) + f_{\theta_2}(\mathbf{x}_i))$ is the softmax probability jointly predicted by $f_{\theta_1}$ and $f_{\theta_2}$, and $\tau_{ps}$ is the confidence threshold for label correction. The average predictions of the two networks could alleviate the confirmation bias of self-training, and achieve a favorable ensemble effect. Further more, the precision of revised labels can be guaranteed with a proper choice of $\tau_{ps}$, as shown in the following Theorem~\ref{th:1}.
\begin{theorem} \cite{yang2021estimating} \label{th:1}
Assume ${\tilde{\alpha}_{c}(\mathbf{x})}$ denotes the conditional probability $\underset{\mathcal{D}_{X}, \mathcal{D}_{\tilde{Y}}}{P}(\tilde{Y} = e_c | {X} = \mathbf{x})$, and ${\alpha_{c}(\mathbf{x})} =
\underset{{\mathcal{D}_{X}, \mathcal{D}_{Y}}}{P}({Y} = e_c | {X} = \mathbf{x})$. $\forall{\mathbf{x}} \sim \mathcal{D}_{\mathbf{x}}$, we have
\begin{equation} \label{eq:th}
    \tilde{\alpha}_{c}(\mathbf{x}) > \frac{1+\rho_{c}}{2} \Longrightarrow \alpha_{c}(\mathbf{x}) > \frac{1}{2}.
\end{equation}
\end{theorem}

Theorem~\ref{th:1} provides us with the guidance of choosing proper $\tau_{ps}$ for label correction. By setting $\tau_{ps}=\frac{1+\rho_{c}}{2}$, we can ensure that the index of the highest prediction is the true class for sample $\mathbf{x}$, as no other class has a higher probability than $\alpha_{c}(\mathbf{x})$.  In practice, however, $\rho_c$ in Eq.~\ref{eq:th} is usually an unknown value, which needs to be estimated. We discuss the problem and study the choice of ReCo hyper-parameters in Sec.~\ref{sec:abl}. The proof of Theorem~\ref{th:1} is provided in the supplementary material.

\subsection{Hybrid Augmentation}
Inspired by~\cite{sohn2020fixmatch}, we seek to enhance the model's generalization and discrimination ability by applying strong augmentation. However, in the proposed framework, the augmentation policy needs to be carefully designed to avoid adversely affecting sample selection or label correction. To this end, we adopt a hybrid weak-strong augmentation strategy for accurate prediction, efficient training, and improving generalization and discrimination. In our experiments, weak augmentation involves flipping and random cropping, and strong augmentation refers to AutoAugment~\cite{cubuk2018autoaugment} or RandAugment~\cite{cubuk2020randaugment} based on different datasets and different noise rates.

In the process of loss modeling and label correction, we simply use raw images (or center crops) for inference and abandon any spatial or color distortion. The goal is to obtain accurate modeling of loss distribution/class probabilities for the best sample selection/label correction. Following~\cite{sohn2020fixmatch} and \cite{nishi2021augmentation}, we apply weak augmentation when performing pseudo labeling on the noisy set $\mathcal{U}$ in Eq.~\ref{eq:cn1}. The pseudo label is then used by its source image as well as another strong augmented view for optimizing the two networks $f_{\theta_1}$ and $f_{\theta_2}$. As a result, consistency is enforced on both weak-weak and weak-strong views. This is different from~\cite{nishi2021augmentation} where the batches for pseudo labeling and optimization are different and disjoint. Our hybrid augmentation strategy could save memory and computation costs while improving generalization with hybrid consistency regularization.

\subsection{Training Objective}
We denote the divided clean and noisy sets after label correction as
\begin{equation} \label{eq:cn2}
\begin{split}
    \hat{\mathcal{X}} & = \{(\mathbf{x}_i, \hat{\mathbf{y}}_i)|(\mathbf{x}_i, \hat{\mathbf{y}}_i) \in \hat{\mathcal{S}}, P_i^{clean} \ge \tau_c\}, \\
    \hat{\mathcal{U}} & = \{(\mathbf{x}_i, p_i)|\mathbf{x}_i \in \hat{\mathcal{S}}_{\mathbf{x}}, P_i^{clean} < \tau_c\},
\end{split}
\end{equation}
and the correspondent mixed sets as 
\begin{equation} \label{eq:mix_cn2}
\begin{split}
    \hat{\mathcal{X}}' = \{ & (l(\mathbf{x}_i, \mathbf{x}_j, \lambda), l(\hat{\mathbf{y}}_i, \mathbf{y}_j, \lambda)) | (\mathbf{x}_i, \hat{\mathbf{y}}_i) \in \hat{\mathcal{X}}, (\mathbf{x}_j, \mathbf{y}_j) \in \hat{\mathcal{X}} \cup \hat{\mathcal{U}}\}, \\
    \hat{\mathcal{U}}' = \{ & (l(\mathbf{x}_i, \mathbf{x}_j, \lambda), l(p_i, \mathbf{y}_j, \lambda)) | (\mathbf{x}_i, p_i) \in \hat{\mathcal{U}}, (\mathbf{x}_j, \mathbf{y}_j) \in \hat{\mathcal{X}} \cup \hat{\mathcal{U}}\}.
\end{split}
\end{equation}
The final training loss of LC-Booster is
\begin{equation}
    \mathcal{L} = 
    \begin{cases}
        \mathcal{L}_{\mathrm{VR}}(\mathcal{X}', \mathcal{U}') + \lambda_r \mathcal{L}_{reg}, \quad \text{before ReCo}, \\
        \mathcal{L}_{\mathrm{VR}}(\hat{\mathcal{X}}', \hat{\mathcal{U}}') + \lambda_r \mathcal{L}_{reg}, \quad \text{after ReCo},
    \end{cases}
\end{equation}

\input{tables/cifar}

where $\mathcal{L}_{\mathrm{VR}}$ is defined in Eq.~\ref{eq: VR1}, $\lambda_r$ is the weight of regularization and 
\begin{equation}
    \mathcal{L}_{reg}=\sum_{c} \pi_{c} \log \left(\pi_{c} \bigg/ \frac{\sum_{\mathbf{x}_i \in \mathcal{X}'_{\mathbf{x}} \cup \mathcal{U}'_{\mathbf{x}}} f_{\theta}^c(\mathbf{x}_i)}
    {|\mathcal{X}'|+|\mathcal{U}'|} \right).
\end{equation}
We apply the same regularization term $\mathcal{L}_{reg}$ as in~\cite{tanaka2018joint, li2020dividemix} to encourage the average output of all samples to be uniform across each class.

%% file: tables/cifar.tex
\begin{table}[ht]
	\vspace{-1em}
	\centering
	\renewcommand{\arraystretch}{0.8}
	\resizebox{0.9\textwidth}{!}{  
	\begin{tabular}{l l | c |c|c|c|c||c|c|c|c} 
		\toprule	 	
			Dataset & &\multicolumn{5}{c||}{CIFAR-10}&\multicolumn{4}{c}{CIFAR-100}\\
			\midrule
			Noise type&  & \multicolumn{4}{c|}{Sym.}& Asym. & \multicolumn{4}{c}{Sym.}\\
			\midrule
			Methods\textbackslash Noise ratio&  & 20\% & 50\%& 80\% & 90\% &40\%&20\% & 50\%& 80\% & 90\% \\
			\midrule
			
			\multirow{2}{*}{Cross-Entropy}	& Best &86.8&79.4	&62.9&42.7&85.0 & 62.0  & 46.7&19.9  &10.1 \\	            
			& Last  & 82.7 &57.9  & 26.1 & 16.8 &72.3  &  61.8 & 37.3& 8.8 & 3.5 \\ 	
			\midrule
			
			\multirow{1}{*}{Mixup}	& 	\multirow{1}{*}{Best} &85.6&87.1	&71.6&52.2&- & 67.8  & 57.3&30.8  &14.6 \\	            
			\multirow{1}{*}{\cite{zhang2017mixup}} & 	\multirow{1}{*}{Last}  & 92.3 &77.3  & 46.7 & 43.9 &-  &  66.0 & 46.6& 17.6 & 8.1\\		 	
			\midrule
			
			\multirow{1}{*}{Bootstrapping}	& 	\multirow{1}{*}{Best} & 91.5 & - & 63.8 & - &91.2 & 69.8  & - & 17.6  & - \\
			\multirow{1}{*}{\cite{reed2014training}} & 	\multirow{1}{*}{Last} & 88.0 & - & 63.4 & - &85.6 & 63.0  & - & 17.0  & - \\	
			\midrule
						
			\multirow{1}{*}{MSLC}	& 	\multirow{1}{*}{Best} & 93.5 & - & 69.9 & - &92.8 & 72.5  & - & 24.3  & - \\	            
			\multirow{1}{*}{\cite{wu2020learning}} & 	\multirow{1}{*}{Last} & 93.4 & - & 68.9 & - &92.5 & 72.0  & - & 20.5  & - \\	
			\midrule
			
			\multirow{1}{*}{M-correction}	& 	\multirow{1}{*}{Best} &94.0&92.0	&86.8&69.1&87.4 &73.9  &66.1 &48.2  &24.3 \\	            
			\multirow{1}{*}{\cite{arazo2019unsupervised}}& 	\multirow{1}{*}{Last}  & 93.8 &91.9  &86.6 &68.7 &86.3  &  73.4 & 65.4& 47.6 & 20.5\\	 						 	
			\midrule
			
			\multirow{1}{*}{Meta-Learning}	& 	\multirow{1}{*}{Best} &92.9&89.3	&77.4&58.7&89.2 &68.5  &59.2 &42.4  &19.5 \\	            
				\multirow{1}{*}{\cite{li2019learning}}& 	\multirow{1}{*}{Last}  & 92.0 &88.8  &76.1 &58.3 &88.6  &  67.7 & 58.0& 40.1 & 14.3\\	 						 	
			\midrule
			
			
			\multirow{1}{*}{DivideMix}	& 	\multirow{1}{*}{Best} &96.1&94.6	&93.2&76.0&93.4 & 77.3  & 74.6&60.2  &31.5 \\	            
				\multirow{1}{*}{\cite{li2020dividemix}}& 	\multirow{1}{*}{Last}  & 95.7 &94.4  & 92.9 & 75.4 &92.1  &  76.9 & 74.2& 59.6 & 31.0\\	 						 	
			\midrule
			
			\multirow{1}{*}{LongReMix}	& 	\multirow{1}{*}{Best} &96.2&95.0	&93.9&82.0&94.7 &77.8  &75.6&62.9  &33.8 \\	            
				\multirow{1}{*}{\cite{cordeiro2021longremix}}& 	\multirow{1}{*}{Last}  & 96.0 &94.7  &93.4 &81.3 &94.3  &  77.5 & 75.1&62.3 & 33.2\\	 						 	
			\midrule
			
			\multirow{1}{*}{DM-AugDesc-WS-WAW}	& 	\multirow{1}{*}{Best} &96.3	&95.4&93.8&91.9 &94.6&79.5  &77.2 &66.4  &41.2 \\	            
				\multirow{1}{*}{\cite{nishi2021augmentation}}& 	\multirow{1}{*}{Last}  & 96.2 &95.1  &93.6 &91.8 &94.3  &  79.2 & 77.0&66.1 & 40.9\\	 						 	
			\midrule
			
			
			\multirow{2}{*}{LC-Booster}	& Best &\textbf{96.4}	&\textbf{95.6}&\textbf{94.7}&\textbf{92.9} &\textbf{95.1}&\textbf{79.6}  &\textbf{77.6}&\textbf{66.9}  &\textbf{48.4} \\	            
			& Last  &\textbf{96.3} &\textbf{95.3}&\textbf{94.4}&\textbf{92.6} &\textbf{95.0}&\textbf{79.5}  &\textbf{77.4}&\textbf{66.5}  &\textbf{48.1} \\				 	
			\midrule
		\bottomrule
    \end{tabular}
    }
	\caption{Comparison with state-of-the-art methods on CIFAR-10 and CIFAR-100 with symmetric (from 20\% to 90\%) and asymmetric noise (40\%). We report both the best test accuracy across all epochs and the averaged test accuracy over the last 10 epochs of training. Results of previous methods are cited from their original papers. \textbf{Bold entries} are best results.}
	\label{tab:cifar}
	\vspace{-1em}
\end{table}			

%% file: sections/5.experiments.tex
\section{Experiments}

\subsection{Datasets and Implementation Details} \label{sec:impl}
We extensively validate the effectiveness of LC-Booster on four noisy-label benchmarks, namely CIFAR-10, CIFAR-100 \cite{krizhevsky2009learning},  Clothing1M \cite{Xiao_2015_CVPR} and WebVision \cite{li2017webvision}.
Clothing1M and WebVision are two large-scale datasets with real-world noisy labels. Clothing1M consists of 1 million training images crawled from online shopping websites and is composed of 14 classes. Labels of Clothing1M are generated from surrounding texts and the overall noise ratio is estimated to be around 40\%. WebVision contains 2.4 million images collected from the Internet, with the same 1000 classes as in ILSVRC12~\cite{deng2009imagenet}. Following previous works~\cite{chen2019understanding, li2020dividemix}, we only use the first 50 classes of the Google image subset for training and test.

For CIFAR-10/100, we experiment with symmetric and asymmetric label noise $\tilde{Y}_{sym}$ and $\tilde{Y}_{asym}$ as described in Sec.~\ref{sec:problem_form}, following the protocol in previous works~\cite{Li_2019_CVPR, li2020dividemix, nishi2021augmentation}. We use an 18-layer PreAct ResNet (PRN18)~\cite{he2016identity} as the network backbone and train it for roughly 300 epochs, following~\cite{nishi2021augmentation}. We adopt SGD as the optimizer with a batch size of 64, a momentum of 0.9, and a weight decay of 0.0005. The initial learning rate is 0.02 and is decayed by a factor of 10 in the middle of training. The warm up period is 10 epochs for CIFAR-10 and 30 epochs for CIFAR-100. As for our method, we perform ReCo at the 100th epoch and set different $\tau_{ps}$ for different noise rates (see the supplementary material). A discussion is also provided in Sec.~\ref{sec:abl} about the choice of the two hyper-parameters.

Following previous baseline methods~\cite{li2020dividemix, cordeiro2021longremix}, we use ImageNet pre-trained ResNet-50 as the backbone for Clothing1M, and use Inception-ResNet v2~\cite{szegedy2017inception} as the backbone for WebVision. More training details and hyper-parameters of the two datasets are delineated in the supplementary material.

\subsection{Comparison with State-of-the-art Methods} \label{sec:comp}
We compare LC-Booster with recent state-of-the-art methods, including Mixup~\cite{zhang2017mixup}, M-correction~\cite{arazo2019unsupervised}, Meta-Learning~\cite{li2019learning}, DivideMix~\cite{li2020dividemix}, LongReMix~\cite{cordeiro2021longremix}, DM-AugDesc~\cite{nishi2021augmentation}. We also compare it with previous label correction methods Bootstrapping~\cite{reed2014training} and MSLC~\cite{wu2020learning}. For fair comparison, we adopt the same backbone as in previous works for all benchmarks.


\input{tables/c1m_webv}

\paragraph{Comparison with synthetic noisy labels.} The results of CIFAR-10/100 are present in Tab.\ref{tab:cifar}. We experiment with different levels of symmetric label noise ranging from 20\% to 90\%, as well as 40\% asymmetric noise for CIFAR-10. Following the metrics in previous works, we report both the best test accuracy across all epochs and the average test accuracy over the last 10 epochs of training. Our LC-Booster outperforms previous state-of-the-art methods across all noise ratios. A remarkable improvement can be seen under the 90\% high noise rate, where 1.0\% and 7.2\% absolute accuracy gains are achieved on CIFAR-10 and CIFAR-100 respectively. This demonstrates the robustness of our method against extreme label noise. Moreover, our method also outperforms previous label correction methods (Bootstrapping, MSLC) by a large margin, which verifies our idea that label correction could be better leveraged with sample selection.

\paragraph{Comparison with real-world noisy labels.} We also validate our method on large-scale noisy labeled data sets. Tab.~\ref{tab:c1m} evaluates LC-Booster on Clothing1M. Our method outperforms previous methods by at least 0.12\% absolute test accuracy. Tab.~\ref{tab:webvision} shows the validation results on (mini) WebVision and ILSVRC12. LC-Booster achieves comparable results on WebVision and state-of-the-art performance on ILSVRC12. These results show that our method can be well applied in real-world scenarios with large-scale data.

\input{tables/abl_module_side}

\input{tables/abl_tau_side}

\subsection{Ablation Studies} \label{sec:abl}
We perform extensive ablation studies to illustrate the effects of our method. For better evaluation, we conduct experiments on CIFAR-10 with 90\% symmetric noise and report the best test accuracy (if not otherwise stated).

\paragraph{Evaluation of ReCo and H-Aug.}
The effect of the two modules are shown in Tab.~\ref{tab:abl_module}. As one can see, the best results are achieved when ReCo and H-Aug. are jointly used, which shows the compatibility of the two modules. Applying either of the two modules individually also brings non-trivial accuracy gain. Moreover, we find that applying ReCo with H-Aug. could obtain a larger improvement than applying ReCo alone (3.1\% vs. 1.4\% on CIFAR-10, 6.6\% vs. 4.8\% on CIFAR-100), which indicates that the advantage of ReCo could be better exploited with a proper augmentation strategy.

\paragraph{Exploring $\tau_{ps}$.} 
$\tau_{ps}$ is defined in Eq.~\ref{eq:S_hat} as the threshold for ReCo. It is proved in Theorem~\ref{th:1} that setting $\tau_{ps} = \frac{1+\rho_{c}}{2}$ guarantees the revised labels are correct. In practice, however, a problem remains that $\rho_{c}$ needs to be estimated given a noisy dataset. Specifically, in the proposed framework, $\rho_{c}$ is still hard to estimate even if the overall noise ratio is determined beforehand (\eg, 90\% symmetric noise). This is because the noise distribution of the clean/noisy set is still unknown and probably changing during training. As this, we simply use $\frac{1+\rho_{c}}{2}$ as the \textit{upper bound} of $\tau_{ps}$, and fine-tune $\tau_{ps}$ from that point for best performance. We present the test accuracy for different $\tau_{ps}$ in Tab.~\ref{tab:tau}, as well as the number of revised samples $|\hat{\mathcal{S}}_{ps}|$ and label precision of $\hat{\mathcal{S}}_{ps}$ for better illustration. One can find that $|\hat{\mathcal{S}}_{ps}|$ decreases monotonically as $\tau_{ps}$ rises. At the same time, the label precision increases and finally arrives at 100\% when $\tau_{ps}$ grows up to 0.95, which is in accordance with Theorem~\ref{th:1} when $\rho_{c}=0.9$. However, even if $\hat{\mathcal{S}}_{ps}$ is absolutely clean when $\tau_{ps}=0.95$,  $|\hat{\mathcal{S}}_{ps}|$ significantly shrinks to less than 500, which is only 1\% of total training data. Such a small $\hat{\mathcal{S}}_{ps}$ can make little change to the total number of clean training samples and could hardly boost model performance. Hence, we discreetly sacrifice the label precision by decreasing $\tau_{ps}$, in exchange for a larger $\hat{\mathcal{S}}_{ps}$. With this sacrifice, a more favorable balance could be achieved between the purity and quantity of revised samples, as shown in Tab.~\ref{tab:tau} that the best accuracy is achieved when $\tau_{ps}=0.8$. Further decreasing $\tau_{ps}$ leads to inferior performance, as more wrongly predicted labels are imbued into $\hat{\mathcal{S}}_{ps}$. More details are available in the supplementary material.

\input{tables/abl_relabel}

\paragraph{Exploring when to perform ReCo.} 
We vary the re-labeling epoch from 50 to 250, with a total of 300 training epochs and $\tau_{ps}=0.8$ as discussed above. As shown in Tab.~\ref{tab:relabel}, the best performance of 92.9\% is achieved at the 100th epoch, which is before the right middle of training. After that, the test accuracy begins to drop. We hypothesize that this is because the model gradually overfits to noisy samples as training progresses, making the predictions less reliable. We also try to perform ReCo multiple times (last column of Tab.~\ref{tab:relabel}, at both the 100th and 200th epoch), which does not bring further accuracy gain. This indicates that re-labeling once is sufficient.

\begin{figure*}[htp]
    \centering
    \subfigure[]
    {\label{fig:retrain}\raisebox{1.4mm}{\includegraphics[width=0.32\textwidth, height = 0.245\textwidth]{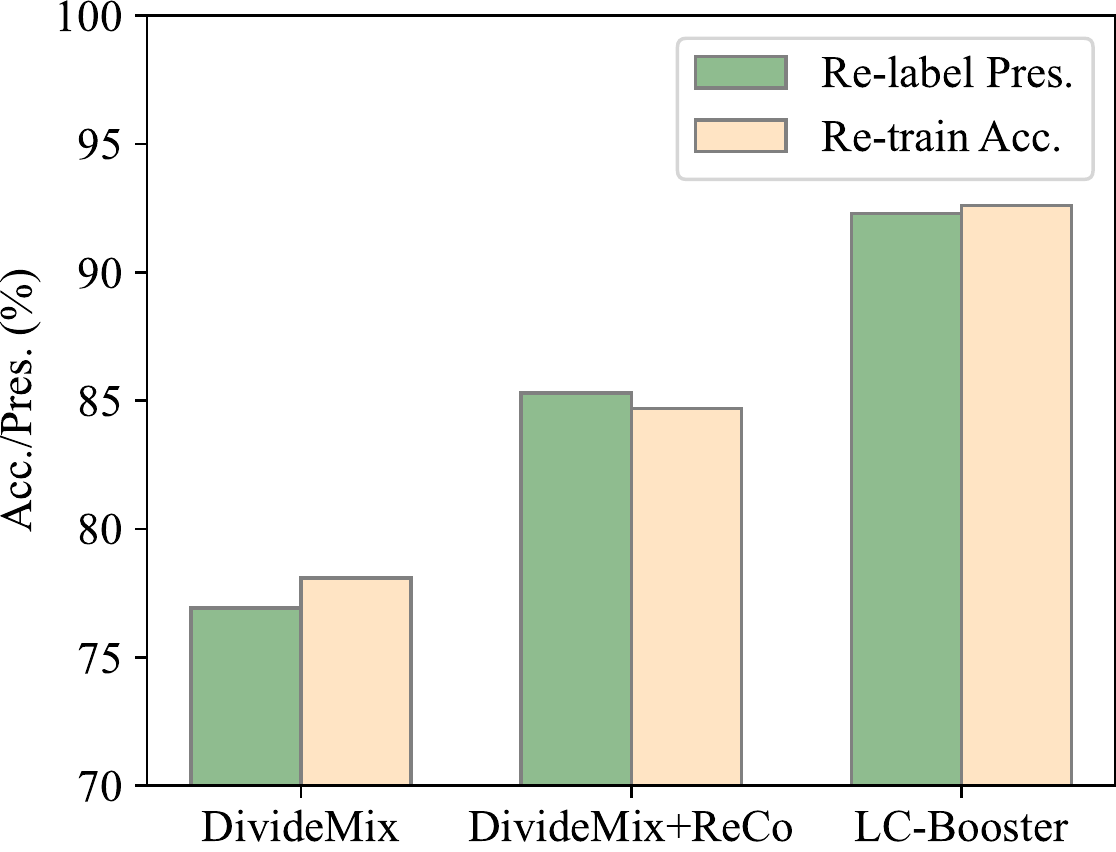}}}
    \subfigure[]
    {\label{fig:testacc}\includegraphics[width=0.32\textwidth, height = 0.25\textwidth]{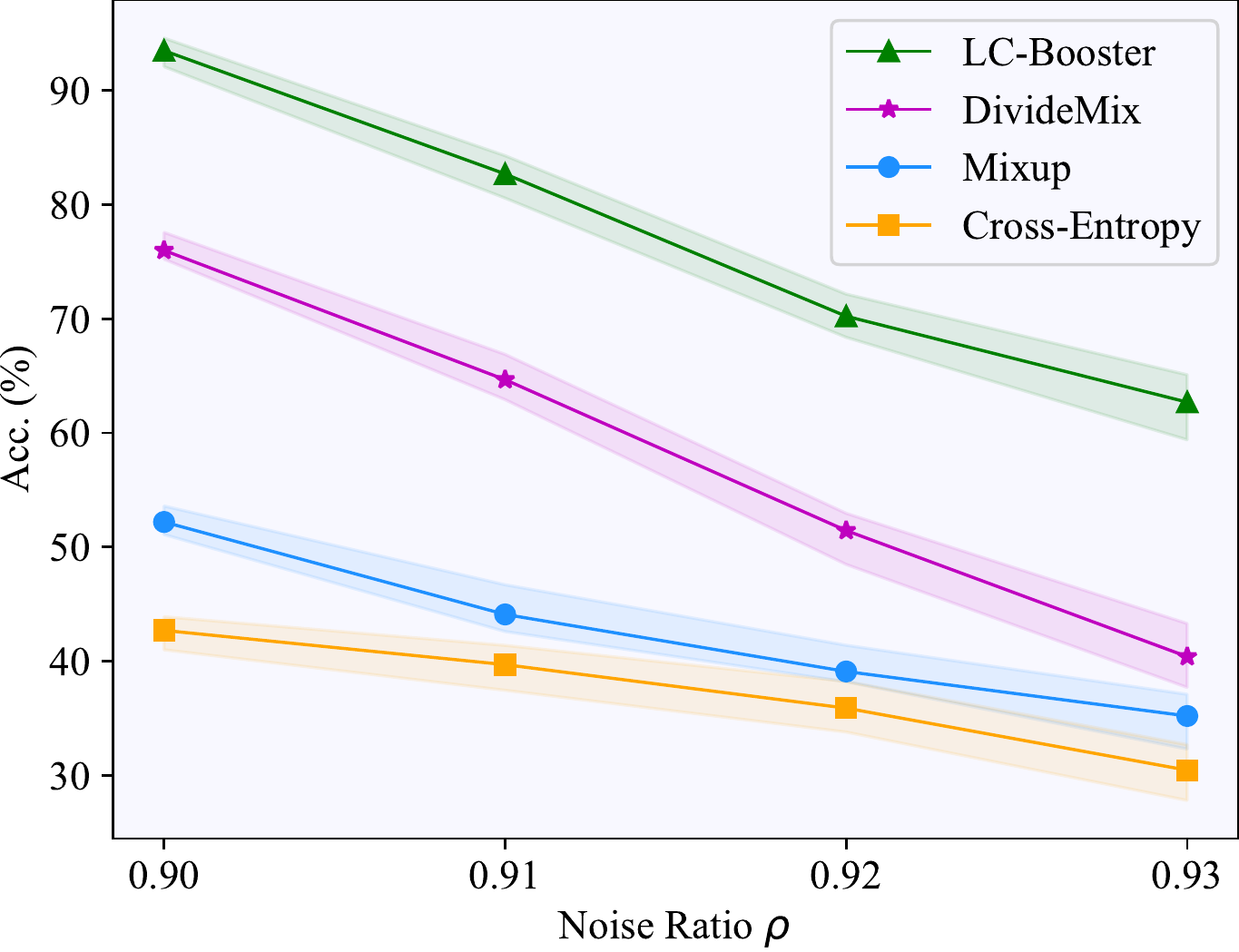}}
    \subfigure[]
    {\label{fig:auc}\raisebox{1.4mm}{\includegraphics[width=0.32\textwidth, height = 0.245\textwidth]{figures/auc.pdf}}}
\vspace{-1em}
\caption{(a) compares the precision of re-labeling (Re-label Pres.) and re-training accuracy (Re-train Acc.) between different methods. Higher indicates stronger cleansing ability. (b) shows the results under even higher extreme label noises, \ie, $\ge$ 90\%. (c) shows the curves of AUC and size of the clean set (\#Clean) on CIFAR-10 with 90\% symmetric noise. Higher AUC indicates that clean samples are selected more precisely based on GMM.}
\end{figure*}

\paragraph{Comparison of re-training performance.}
Here, we compare the re-labeling quality of LC-Booster with other methods. We first re-label the noisy dataset with the trained model. Then, a randomly initialized PRN18 is trained from scratch using re-labeled samples. We compare both the precision of new labels and test accuracy of re-trained models in Fig.~\ref{fig:retrain}. It can be seen in the figure that our method achieves the highest re-labeling precision and re-training accuracy. Remarkably, the re-labeling precision achieves over 90\% under 90\% symmetric noise, demonstrating the superior cleansing ability of our method. Moreover, simply applying ReCo with DivideMix could also obtain a higher re-labeling precision as well as re-training accuracy.

\paragraph{Evaluation under extreme label noises.} 
We evaluate the robustness of our method under even higher extreme label noises, \ie, $\ge$ 90\% symmetric noise. To the best of our knowledge, no previous attempts have been made under such heavy label noise. The results are shown in Fig.~\ref{fig:testacc}. LC-Booster consistently achieves the best results across all noise rates. Furthermore, it can also be observed that the performance gap between LC-Booster and DivideMix increases as the noise rate grows from 90\% to 93\%. This demonstrates the superior robustness of our method under extreme label noise. 

\subsection{Visualization} \label{sec:vis}
\begin{figure}[htp]

\centering
\begin{minipage}{.56\textwidth}
    \subfigure[DivideMix]
    {{\includegraphics[width=0.46\textwidth, trim=0cm 0cm 0.5cm 0.3cm, clip]{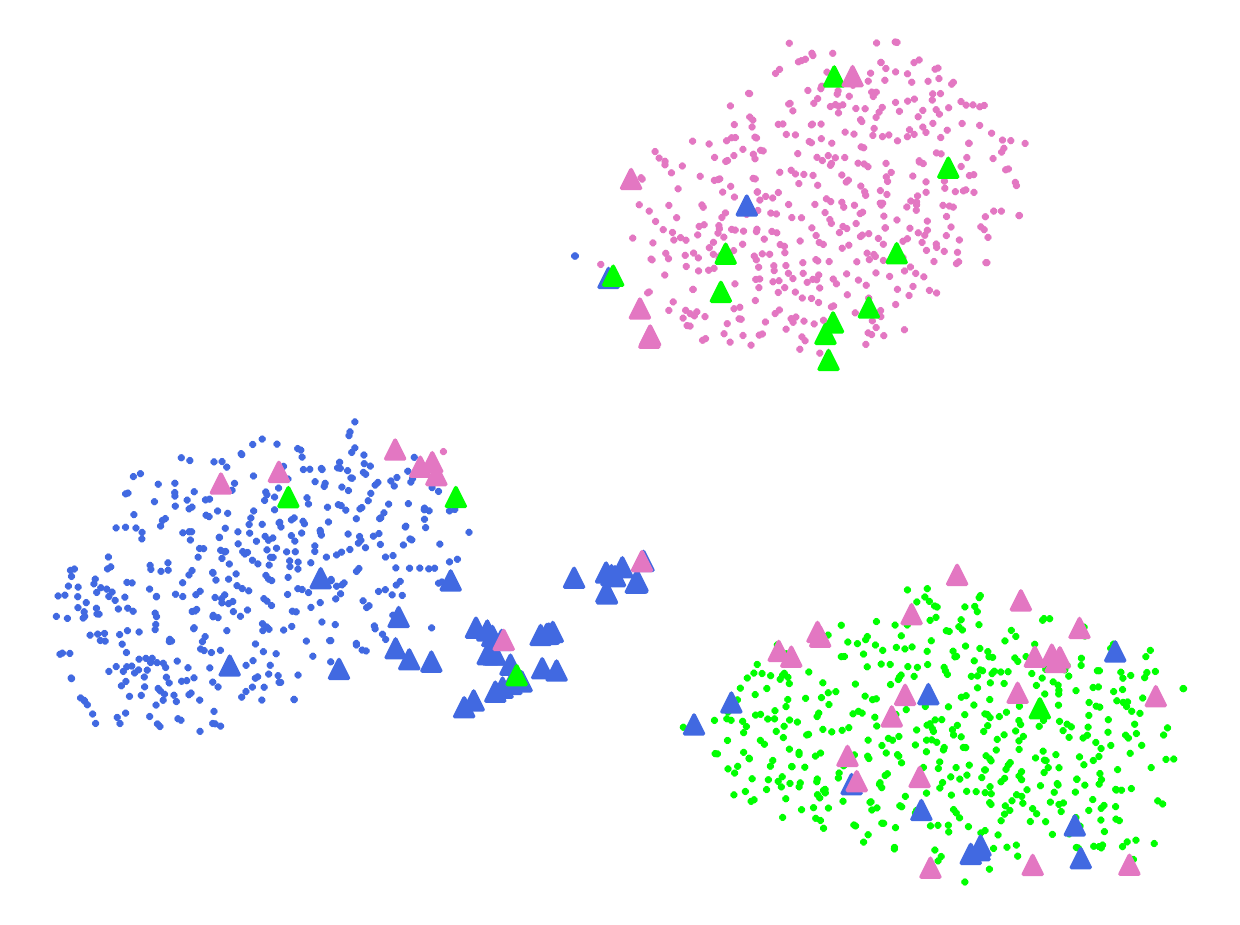}}}
    \subfigure[LC-Booster]
    {{\includegraphics[width=0.46\textwidth, trim=0cm 0cm 0.8cm 0.3cm, clip]{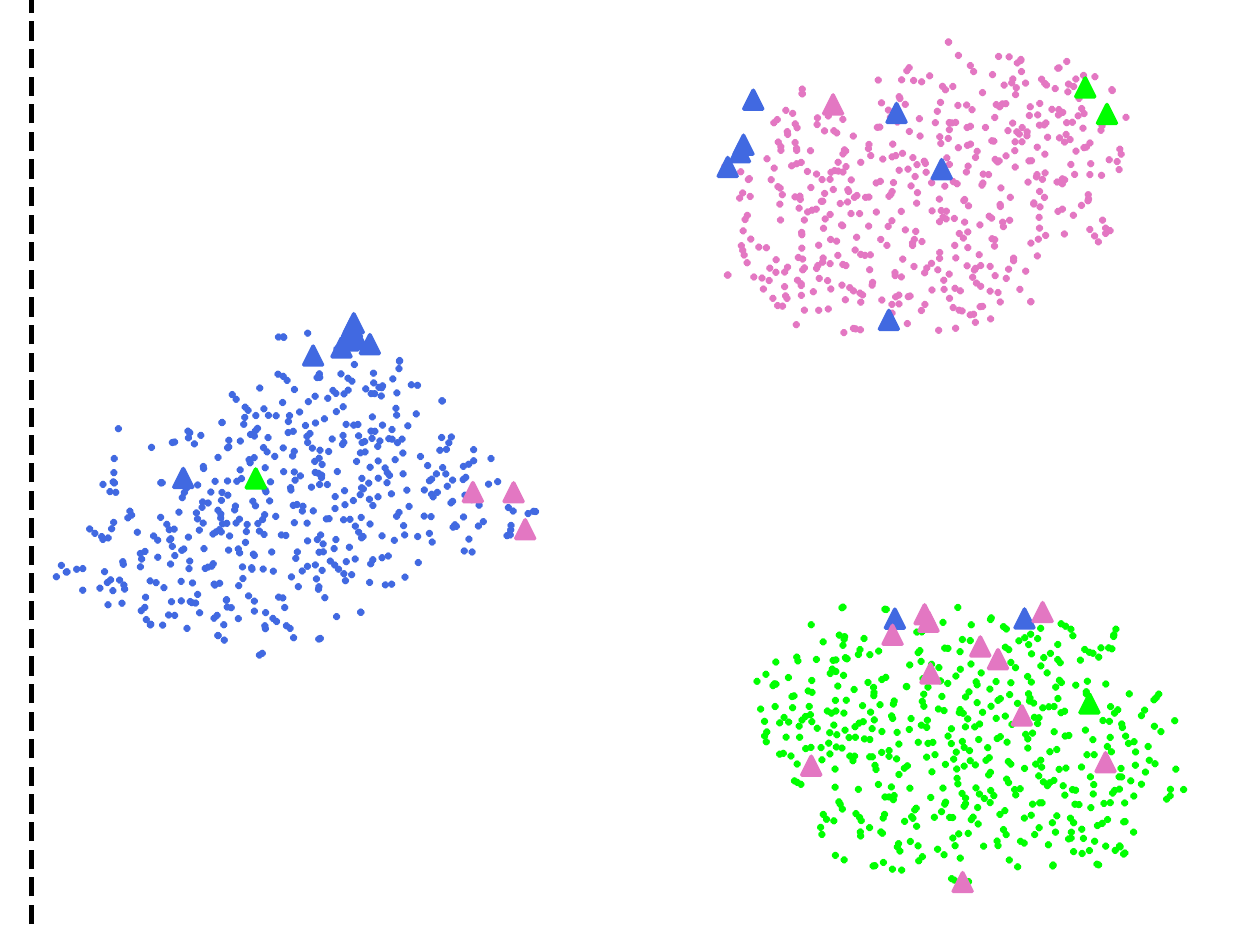}}}
\end{minipage}
\begin{minipage}{.4\textwidth}
    \caption{Visualization of embedded features on CIFAR-10 with 90\% symmetric noise. Three colors indicate the first three classes of CIFAR-10. Correct predictions are marked as dots and false predictions as triangles. Best viewed in color. \label{fig:tsne}}
\end{minipage}

\end{figure}

\paragraph{Learned embeddings.}
We compare feature distributions of DivideMix and our LC-Booster using t-SNE in Fig.~\ref{fig:tsne}. For explicitness, we only visualize the first three classes of CIFAR-10 with 90\% symmetric noise. A complete distribution of 10 classes is provided in the supplementary material. One can see that there exist some obvious outliers of DivideMix, while features of our method are better clustered. Moreover, LC-Booster has fewer false predictions (marked as triangles) compared with DivideMix, demonstrating its robustness under a high noise ratio.

\paragraph{AUC and size of clean set.}
We show the dynamics of AUC and the size of the clean set in Fig.~\ref{fig:auc}. Numbers are from experiments on CIFAR-10 with 90\% symmetric noise. We use the clean probabilities output by GMM for calculating AUC. As shown in the figure, LC-Booster consistently achieves higher AUC than DivideMix during training, which shows that our method is able to select clean samples more precisely. Moreover, after the 100th epoch of performing ReCo, the size of the clean set in LC-Booster significantly rises and surpasses that of DivideMix by a large margin. The effective expansion of the clean set helps to explain the superior performance of our method. More curves of performing ReCo at different epochs are shown in the supplementary material.

%% file: tables/c1m_webv.tex
\begin{table}[h]

\begin{minipage}{0.41\linewidth}
\centering
\small
\resizebox{\linewidth}{0.47\linewidth}{  
	\begin{tabular}	{l|c}
		\toprule	 	
			Method  & Test Accuracy \\
			\midrule			
			Cross-Entropy & 69.21 \\
			F-correction~\cite{patrini2017making}  &69.84\\		
			M-correction~\cite{arazo2019unsupervised}&  71.00 \\			
			Joint-Optim~\cite{tanaka2018joint}  & 72.16\\		
			

			Meta-Learning~\cite{li2019learning}  & 73.47\\				
			P-correction~\cite{yi2019probabilistic} & 73.49 \\
			DivideMix ~\cite{li2020dividemix} & 74.76\\
			ELR+ ~\cite{liu2020early} &74.81\\
			LongReMix ~\cite{cordeiro2021longremix} &74.38\\
			DM-AugDesc ~\cite{nishi2021augmentation} &75.11 \\
			\midrule
			Ours & \textbf{75.23}\\
		\bottomrule
	\end{tabular}
}
\vspace{0.2em}
	\caption{Comparison with state-of-the-art methods in test accuracy (\%) on Clothing1M. Results for baseline methods are cited from original papers. The best entry is marked in \textbf{bold}.}
	\label{tab:c1m}
\end{minipage}\hspace{3mm}
\begin{minipage}{0.55\linewidth}
\centering
\small
\resizebox{\linewidth}{0.32\linewidth}{
	\begin{tabular}	{l|c|c|c|c}
		\toprule	 	
			\multirow{2}{*}{Method}  & \multicolumn{2}{c|}{WebVision} & \multicolumn{2}{c}{ILSVRC12}\\
			\cmidrule{2-5}
			& Top1 & Top5& Top1 & Top5\\
			\midrule			
			F-correction~\cite{patrini2017making} & 61.12 & 82.68&  57.36 &82.36\\		
			Decoupling~\cite{malach2017decoupling} & 62.54 &84.74 & 58.26 &82.26\\
			MentorNet~\cite{jiang2018mentornet} & 63.00 & 81.40 & 57.80 & 79.92 \\
			Co-teaching~\cite{han2018co} & 63.58 & 85.20 & 61.48 & 84.70 \\
			Iterative-CV~\cite{chen2019understanding} & 65.24 &85.34 & 61.60 & 84.98 \\
			DivideMix~\cite{li2020dividemix} & 77.32 & 91.64 & 75.20 & 90.84\\
			LongReMix \cite{cordeiro2021longremix} & 78.92 & \textbf{92.32} & - & - \\
            NGC \cite{wu2021ngc} & \textbf{79.16} & 91.84 & 74.44 & 91.04 \\
			\midrule
			Ours & 78.29 & 92.18 & \textbf{75.44} & \textbf{91.26} \\
		\bottomrule
	\end{tabular}
}
\vspace{0.2em}
	\caption{Comparison with state-of-the-art methods trained on (mini) WebVision dataset. Numbers denote top-1 (top-5) accuracy (\%) on the WebVision validation set and the ILSVRC12 validation set. Results for baseline methods are cited from their original papers. \textbf{Bold entries} are best results.}
	\label{tab:webvision}
\end{minipage}

\end{table}

%% file: tables/abl_module_side.tex
\begin{SCtable}[][t]\setlength{\tabcolsep}{5pt}
\centering
\scalebox{0.85}{
    \begin{tabular}{c|c|c|c}
        \toprule
         ReCo & H-Aug. & CIFAR-10 & CIFAR-100 \\
         \hline
         && 83.9 & 31.5 \\
         \checkmark & &85.3 & 36.3 \\
         & \checkmark &89.8 & 41.8 \\
         \checkmark &\checkmark &\textbf{92.9}&\textbf{48.4} \\
         \bottomrule
    \end{tabular}
}
\caption{Evaluation of ReCo and H-Aug. in the proposed framework. The noise type is 90\% symmetric noise for both CIFAR-10 and CIFAR-100. \textbf{Bold entries} are best results.}
\label{tab:abl_module}
\end{SCtable}

%% file: tables/abl_tau_side.tex
\begin{SCtable}[][t]\setlength{\tabcolsep}{5pt}
\centering
\scalebox{0.98}{
    \begin{tabular}{c|c|c|c|c|c}
        \toprule
         $\tau_{ps}$ & 0.0 & 0.5  & 0.7 & 0.8 & 0.95 \\
         \hline
         $|\hat{\mathcal{S}}_{ps}|$ & 50000 & 42641 & 29715 & 13405 & 483 \\
         Label Pres. (\%) & 84.6 & 91.5 & 97.4 & 99.1 & 100.0 \\
         Test Acc. (\%) & 84.9 & 90.5 & 92.8 & \textbf{92.9} & 91.3 \\
         \bottomrule
    \end{tabular}
}
\caption{Exploration of different $\tau_{ps}$. $\hat{\mathcal{S}}_{ps}$ is defined in Eq.~\ref{eq:S_hat} as the set of revised samples. Label Pres. denotes the label precision of $\hat{\mathcal{S}}_{ps}$. The best result is marked in \textbf{bold}.}
\label{tab:tau}
\end{SCtable}

%% file: tables/abl_relabel.tex

\begin{table}[h]
    \centering
    \begin{tabular}{c|c|c|c|c|c|c}
        \toprule
        Epoch(s) & 50 & 100 & 150 & 200 & 250 & 100\&200 \\
        \hline
        Test Acc. (\%) & 91.9 & \textbf{92.9} & 91.5 & 90.3 & 89.7 & 92.7 \\
        \bottomrule
    \end{tabular}
    \caption{Exploring which epoch(s) to perform ReCo. The last column means re-labeling twice at the 100th and 200th epoch. The best result is marked in \textbf{bold}.}
    \label{tab:relabel}
    \vspace{-1em}
\end{table}

%% file: sections/6.conclusion.tex
\vspace{-1em}
\section{Conclusion}

In this paper, we propose LC-Booster, a novel framework for learning with extremely noisy labels. LC-Booster naturally leverages label correction with sample selection, to make a larger and more purified clean set that effectively boosts model performance. Through extensive experiments on multiple benchmarks, we show that LC-Booster consistently demonstrates superior performance compared with state-of-the-art methods. We hope the proposed learning paradigm could inspire future research along this direction for the problem of LNL.

\textbf{Acknowledge.} This research is supported by the National Research Foundation, Singapore under its AI Singapore Programme (AISG Award No: AISG2-PhD-2021-08-008), 
NUS Faculty Research Committee Grant (WBS: A-0009440-00-00), and the EPSRC programme grant Visual AI EP/T028572/1. We thank Google TFRC for supporting us to get access to the Cloud TPUs. We thank CSCS (Swiss National Supercomputing Centre) for supporting us to get access to the Piz Daint supercomputer. We thank TACC (Texas Advanced Computing Center) for supporting us to get access to the Longhorn supercomputer and the Frontera supercomputer. We thank LuxProvide (Luxembourg national supercomputer HPC organization) for supporting us to get access to the MeluXina supercomputer.


%% file: tables/tau_ps.tex

\begin{table*}[ht]
	\centering
	\renewcommand{\arraystretch}{0.8}
	\resizebox{0.9\textwidth}{!}{  
	\begin{tabular}{c| c|c|c|c|c || c|c|c|c} 
		\toprule	 	
			Dataset &\multicolumn{5}{c||}{CIFAR-10}&\multicolumn{4}{c}{CIFAR-100}\\
			\midrule
			Noise type& \multicolumn{4}{c|}{Sym.}& Asym. & \multicolumn{4}{c}{Sym.}\\
			\midrule
			Hyper-parameters\textbackslash Noise ratio & 20\% & 50\%& 80\% & 90\% &40\%&20\% & 50\%& 80\% & 90\% \\
			\midrule			
			$\tau_{ps}$	& 0.9 & 0.9	& 0.8 & 0.8 & 0.8 & 0.8  & 0.8 & 0.5 & 0.3 \\
			\midrule
			$\lambda_{u}$ & 0 & 25 & 25 & 50 & 0 & 25 & 150 & 150 & 150 \\
			\midrule
		\bottomrule
    \end{tabular}
    }
	\caption{Values of $\tau_{ps}$ and $\lambda_{u}$ for different noise types and noise rates.}
	\label{tab:tau_ps}
\end{table*}